# A Close Look into Human Activity Recognition Models using Deep Learning

Wei Zhong Tee
Undergraduate Computer Science Major
University of Wisconsin-Eau Claire
Eau Claire, United States
teewz1076@uwec.edu

Rushit Dave
Assistant Professor, Computer Science Dept
University of Wisconsin-Eau Claire
Eau Claire, United States
daver@uwec.edu

Jim Seliya
Assistant Professor, Computer Science Dept
University of Wisconsin-Eau Claire
Eau Claire, United States
seliyana@uwec.edu

Mounika Vanamala
Assistant Professor, Computer Science Dept
University of Wisconsin-Eau Claire
Eau Claire, United States
vanamalm@uwec.edu

*Abstract*— Human activity recognition using deep learning techniques has become increasing popular because of its high effectivity with recognizing complex tasks, as well as being relatively low in costs compared to more traditional machine learning techniques. This paper surveys some state-of-the-art human activity recognition models that are based on deep learning architecture and has layers containing Convolution Neural Networks (CNN), Long Short-Term Memory (LSTM), or a mix of more than one type for a hybrid system. The analysis outlines how the models are implemented to maximize its effectivity and some of the potential limitations it faces.

## I. Introduction

Human activity recognition (HAR) has been a popular topic of research [1] for its high usability in fields such as the medical industry. Its ability to be used in health care systems [2] has made it increasingly popular, aiding physicians in making better decisions and allowing better allocation of medical resources based on automated monitoring. HAR's reach has also gone beyond the medical industry and is used by people just to monitor [3] their physical activities or can also be used to detect anomalous events among the elderly [4], such as falls. Furthermore, as human exercise is important for people of all ages, but especially for the elderly, there is a need to be able to monitor such events frequently to track fitness, human capabilities and for detecting anomalous events. The use of more traditional machine learning algorithms has become less popular as more efficient and capable deep learning methodologies are developed. Popular usages of deep learning algorithms include Convolution Neural Networks (CNN), Long Short-Term Memory (LSTM), Recurrent Neural Networks (RNN) and more.

## II. Background

Exploring how different data collection, data preprocessing and deep learning algorithms and effect a model's capability to recognize activities is important for determining important factors [6] and conserving resources associated with the implemented models. Human activity recognition is often divided into two categories: vision-based recognition [7] and sensor-based recognition [8]. As the name suggests, vision-based recognition models utilize one or more cameras for collecting video samples of human activities. Work with vision-based models include using multiple views for determining a single action [9] or a single view, treating the video sample as a human silhouette for making predictions [10]. Sensor-based recognition is a far larger area of research [11] and application as sensors from mobile devices or body attachments are far more obtainable and efficient than using cameras for both training and real-world applications [12]. Some popular datasets that are used for models include the OPPORTUNITY [13], Skoda Checkpoint [14], UCI-HAR [15], WISDM [16], MHEALTH [17] and PAMAP2 [18] datasets. Within the world of human activity recognition, some of the most popular machine learning algorithms [19] have been identified for its high accuracy and robustness across different datasets. Such algorithms include the Support Vector Machine (SVM), k-Nearest Neighbor (kNN) and Random Forest classifier (RF) [20]. However, as proper data preprocessing feature extraction, training and classification methods were developed using deep learning techniques, machine learning [21] methods fell out of favor due to the much more resource efficient and capable [22] nature of deep learning algorithms [23]. This survey investigates some state-of-the-art human activity recognition models that are built using deep learning methodologies based on CNN, LSTM and hybrid layers within the model's architecture.

## III. Human Activity Recognition Using Deep Learning Methodologies

This section presents some featured studies that propose models based on CNN, LSTM and hybrid deep learning architectures.

### A. CNN

A recent study [24] that utilized a lightweight CNN model for human activity recognition based on wearable devices. It utilized different datasets that obtained data from smartphones of portable sensors. Such datasets include the UCI-HAR, OPPORTUNITY, UNIMIB-SHAR, PAMAP2 and WISDM dataset. Additional studies that utilized data from smartphones along with CNN based human activity recognition models can be found in [25] and [26], where the models performed well with such data.

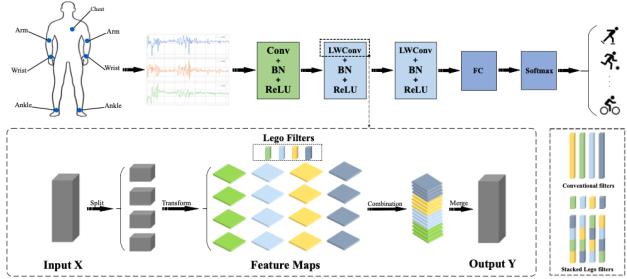

Figure 1. The architecture of a lightweight CNN based model featured in [24]. The model is unique for having a set of lower-dimensional filters which is used as Lego bricks and are stacked for conventional filters resulting in the model being independent of any form of special network structures.

The model is unique for having a set of lower-dimensional filters which is used as Lego bricks and are stacked for conventional filters resulting in the model being independent of any form of special network structures. Signals from the data are split into windows of fixed sizes and overlapping between adjacent windows is tolerated to maintain the continuity of the activities. Ordinary convolution filters that are replaced by the lower dimensional Lego filters makes the model extremely efficient and the Lego filters are simultaneously optimized during the training stage. A straight through estimator (STE) is used as the binary mask for optimizations. To exploit the intermediate feature maps and accelerate convolutions, a classical split transform merge three stage strategy is utilized. ReLU activation functions are used for the different convolution layers and lastly a Softmax function is used to make the activity label prediction based on the network's final output. The local loss functions are implemented using a cross entropy between the prediction of the local liner classifier and its target. The other loss function is based on a kernel of size three by three and a stride and padding of one. The Lego CNN reduces memory and computation costs compared to normal CNN with far lower training parameters and achieved high accuracy through training using the local loss functions. To put the model and datasets to the test, the model was trained on all the listed datasets and evaluated based on overall classification accuracy and the weighted F1 score. As a result, the. model performed best with the UCI-HAR and WISDM datasets, getting an accuracy of 96.9% and 98.82% and F1 scores of 96.27% and 97.51% respectively.

Another study [27] that proposed a model based on a CNN was one that was focused on unobtrusive activity recognition of elderly people using anonymous binary sensors. A similar study used a CNN based model that is trained using novel spectral data strategies is presented with the goal of detecting freezing of gait in Parkinson's disease patients [28]. The model from [27] was built using a deep convolution network (DCNN) using an Aruba annotated open dataset that could predict 10 activities that was recorded by a single elderly woman over the span of eight months.

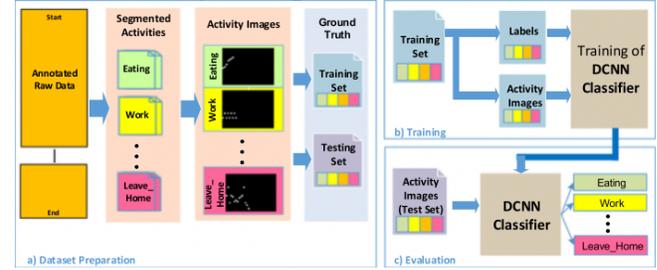

Figure 2. This figure shows the experimental setup [27] used for its DCNN classifier. It focused on using data collected from elderly people using anonymous binary sensors.

As mentioned earlier, it is common for data to be collected using specialized body worn sensors [29], as models proved effective with such techniques in [30]. To preprocess the raw data for the model, the activities are segmented based on their occurrences in the dataset, put through a sliding window to obtain fragmented samples and then lastly the activity was converted into an activity image. The activity image was a simple black and white binary images that had on and off signals that were black and white to represent the activity. The DCNN classifier consisted of three convolution layers, followed by pooling layers that was used for feature extraction. The output of the final max-pooling layer was then flattened and fed to the neurons of the connected layers and in the end the final layer is linked to 10 outputs. The predictive ability of activity was evaluated on accuracy, precision, recall, F1 score and error rate. The activity that had the highest accuracy was going from the bed to the toilet, followed by working, with accuracy scores of 99.99% and 99.85% respectively. On average the DCNN model was able to achieve an accuracy of 98.54% when predicting 10 activities, and 99.23% when predicting 8 activities.

*B. LSTM*

Bi-directional LSTM (BiLSTM) techniques in human activity recognition have become increasingly popular because of its effectivity with extracting features and making predictions. Models from studies [31] and [32] utilize a BiLSTM for the model's predictive capabilities, while [33] employs it for unique feature extraction techniques. The model utilizes the residual block to extract spatial features from multidimensional signals of MEMS inertial sensors. A CNN based architecture is used in the residual block, as it can extract local spacial features automatically. To get the spatial features from different sensor signals, a 2D CNN residual network with 23 kernels of size 2 x 2 is used. The stride length is 2 and a batch normalization layer is added to speed up training and avoid issues of covariate shift. A ReLU activation function is then used before it goes through another same convolution layer with the same setup except it has a stride of 1. Data was standardized using before being fed into the model. The convolution kernels had dimensions of 2 x 2 because it had the best recognition accuracy. 32 convolution kernels were selected for a balance between the model size and training cost. Cross entropy of the model was minimized using the ADAM optimizer with an optimal learning rate of 0.0003, 0.0006 and 0.00003. A batch side of 64 was used and models were trained a total of 80 times. Forward and backward dependencies of the features are then used by the BiLSTM layer and then features are fed into a Softmax layer for HAR. The model was put to the test one three different datasets: a homemade one, WISDM and PAMAP2. The proposed models also had significantly fewer training parameters and even more ideal results than other deep learning-based models. It performed well for all three datasets, with the homemade dataset getting an accuracy of 96.95%, WISDM getting 97.32% and the PAMAP2 getting 97.15%.

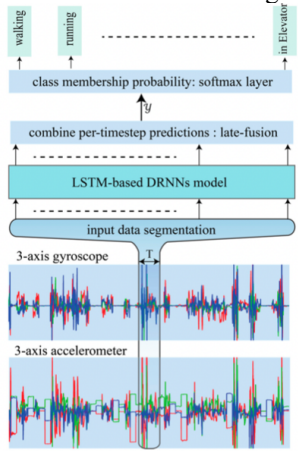

Figure 3. The LSTM based model proposed in [34] that utilizes accelerometer and gyroscope data.

A unidirectional, bidirectional, and cascaded architectures based on LSTM was presented in [34]. The model uses a deep recurrent neural network for capturing long-range dependencies in variable-length input sequences. This means that it can classify variable-length windows of human activities. 3-axis accelerometer and gyroscope data were used and segmented in time series windows and fed into the model. Upon input, the model outputs a sequence of scores that represents activity labels in which there is a label prediction for the varying time steps. A vector of scores represents the prediction. The SoftMax layer is then used to convert prediction scores into probabilities. Within the LSTM based classifier, three different configurations were built to test three different models. A unidirectional LSTM based DRNN model, a bidirectional LSTM-based DRNN model and lastly a Cascaded Bidirectional and Unidirectional LSTM-based DRNN model. Several datasets were used in testing the effectiveness of the models: the UCI-HAD dataset, the USC-HARD dataset, the Opportunity dataset, the Daphnet FOG dataset and lastly the Skoda dataset. Each model was trained using 80% of the data and the remaining 20% was used for validation. The weights of the model were random initially then constantly updated by the mean cross entropy between the ground truth labels and predicted output labels. The Adam optimizer was used for minimizing the cost of backpropagating gradients and updating model parameters. Overall, the unidirectional LSTM based DRNN performed best with the USC-HAD dataset. It had an overall accuracy of 97.8% and an average precision of 97.4%. Furthermore, the hybrid model performed better than traditional machine learning algorithms [36] such as KNN and stand-alone deep learning algorithm-based models such as CNN. Models from [37] and [38] also employ similar techniques of using LSTM layers within the context of an RNN based model for human activity recognition [35].

## C. Hybrid Models

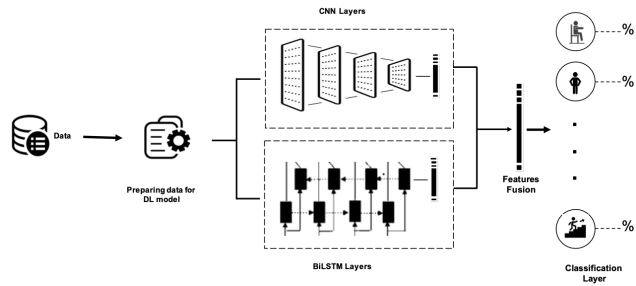

Figure 4. : A hybrid model that utilizes both CNN and (Bi)LSTM layers from [39].

The purpose of another study [39] was to introduce a model for human activity recognition that uses a CNN with varying kernel dimensions that work with a bi-directional long short-term memory (BiLSTM) layer to capture features at different resolutions. This is like what Is seen in proposed models from studies [40], [41] and [42]. These studies all use the combination of both CNN and LSTM layers in the model for advance feature extraction and processing techniques that make the classification ability of the model more robust. The model from [43] excels at extracting spatial and temporal features from the sensors using the CNN and BiLSTM. The model's input had to be transformed into the output value using an activation function. The CNN then performed dimensionality reduction of the input data using both maximum and average pooling. The LSTM layer involves automatic learning of high-level features that are related to long term ways across time steps. The BiLSTM part was used for obtaining temporal representation about activity recognition and can access context in forward and backwards directions. The WISDM and UCI datasets were used, and the model was fitted on 30 epochs and had a batch size of 128 samples. The model's accuracy on the WISDM dataset was 98.53% and the UCI dataset was 97.05%.

A total of four models were presented in [45]; the four models include a convolution neural network (CNN) with a Gated Recurrent Unit (GRU), a CNN with a GRU and attention, a CNN with a GRU and a second CNN, and a CNN with Long Short-Term Memory (LSTM) and a second CNN. The study proposes four different human activity recognition models based on deep learning that utilizes channel state information (CSI) in WiFi 902.11n. A public dataset was utilized. Data was collected using the Linux CSI 802.11n tool. This tool is ideal for describing WiFi signals. Actions that occur between a WiFi transmitter and receiver are recorded, and channels display different amplitudes for different actions. The activities that were labeled includes: Lying down, Fall, Walk, Run, Sit down and Stand up. Six people performed each activity 20 times in an indoor office and the data was split into 80% and 90% groups for training and testing. The best performing model was the CNN-GRU model. The CNN-GRU model consists of three parts: input, features extraction and classification. For input, the CSI data is reaped to be suitable inputs to CNNs, so their matrices are reshaped into 1000 x 30 x 3. For feature extraction, two convolutional layers and a GRU layers are used. The input data is put through filters with a size 5 x 5 x 128 kernel and size 1 x 1 stride followed by batch normalization, ReLU activation, average pooling, and dropout with a value of 0.6 is used. Output passes through a flattening layer with time distributed input to convert the data into a vector suitable for the GRU layer. It obtained an accuracy of 99.46%, precision of 99.52% and AUC of 99.90%, outperforming the other three models and other state-of-the-art models. A similar study [46] utilized a deep-neural network-based model that is trained by utilizing transfer learning and shared-weight techniques to classify human activity from cameras. The model which contained Pre-trained CNN and Shared-Weight LSTMRes layers obtained an accuracy of 97.22%.

IV. A SUMMARY AND ANALYSIS OF FEATURED WORKS

| Title | DL based Methodologies | Experimental Results | Limitations |
|---|---|---|---|
| Layer-wise Training Convolutional Neural Networks with Smaller Filters for Human Activity Recognition Using Wearable Sensors [24] | A lightweight CNN model for human activity recognition based on wearable devices. The model consists of a set of lower-dimensional filters which is used as Lego bricks and are stacked for conventional filters. | The model performed best with the UCI-HAR and WISDM datasets, getting an accuracy of 96.9% and 98.82% and F1 scores of 96.27% and 97.51%. | The use of smaller Lego filters results in a slight decrease in performance compared to a model based on a traditional CNN. |
| Unobtrusive Activity Recognition of Elderly People Living Alone Using Anonymous Binary Sensors and DCNN [27] | A model based on a deep convolution network (DCNN) and focused on unobtrusive activity recognition of elderly people using anonymous binary sensors. | On average the DCNN model was able to achieve an accuracy of 98.54% when predicting 10 activities, and 99.23% when predicting 8 activities. | Model's dataset was limited to working with a high-cost setup of binary sensors that recorded the movements for training and validation of the model. Untested with more accessible devices such as smartphones or smartwatches. |
| Human Activity Recognition Based on Residual Network and BiLSTM [33] | Utilized a residual block for extract spatial features from multidimensional signals and bi-directional LSTM (BiLSTM). | It performed well for all three datasets used, with the homemade dataset getting an accuracy of 96.95%, WISDM getting 97.32% and the PAMAP2 getting 97.15%. | Specific activity labels from the datasets that are unbalanced result in lower predicting success. Future work intends to address this to further increase accuracy. |
| Deep Recurrent Neural Networks for Human Activity Recognition [34] | A unidirectional, bidirectional, and cascaded architectures based on LSTM that uses a deep recurrent neural network for capturing long-range dependencies in variable-length input sequences. | The unidirectional LSTM based DRNN performed best with the USC-HAD dataset. It had an overall accuracy of 97.8% and an average precision of 97.4%. | The model's capabilities were only tested with basic human activities on a small scale and not tested with large scale complex activities. |
| Sensor-Based Human Activity Recognition with Spatio-Temporal Deep Learning [38] | This model uses a convolution neural network (CNN) with varying kernel dimensions that work with a bi-directional long short-term memory (BiLSTM) layer to capture features at different resolutions. | The model performed well for three datasets, with the homemade dataset getting an accuracy of 96.95%, WISDM getting 97.32% and the PAMAP2 getting 97.15%. | Power and memory usage was not considered for this model, hence the potential for low-powered devices to struggle with such a model. |
| Utilizing deep learning models in CSI-based human activity recognition [44] | The study proposes four different human activity recognition models based on deep learning that utilizes channel state information (CSI) in WiFi 902.11n. | The best performing model was the CNN-GRU model. It obtained an accuracy of 99.46%, precision of 99.52% and AUC of 99.90%, outperforming the other three models. | No use of denoising in the signals from the data before training. |

a. Featured Studies Using Deep Learning Methodologies for Human Activity Recognition

Table 1 summarizes the features studies that were surveyed as state-of-the-art human activity recognition models that utilize deep learning-based architecture to achieve its predictive capabilities. The papers include a division of models that utilize CNN layers, LSTM layers and hybrid models that utilize more than one algorithm, such as employing both CNN and LSTM layers in the model.

## V. LIMITATIONS

The papers that were analyzed in depth had limitations that were very specific to the unique architecture of the models. General limitations with human activity recognition with deep learning, while being somewhat negligible, is that often datasets must be very large before a model can become efficient with recognizing complex activities. Furthermore, models that deal with complex activities with many points have many complex steps of data pre-processing to balance out the data or shift weights to allow for overall accuracy of recognition for different activities. Most model's limitations do not come from the deep learning classifiers but using feature extraction and data preprocessing. Most state-of-the-art models even implement entire deep learning based architectural layers for the feature extraction and preprocessing process.

## VI. LIMITATIONS

This article investigates the technical details of the start-of-the-art human activity recognition models that are built using deep learning layers. A couple of CNN, LSTM and hybrid-based architectures are features in this article along with its setup features, experimental results, and limitations. As models are continuously developed on the back of previously completed research and implementations, the limitations that models face become more and more insignificant to general real-world performance and usage. Such a conclusion can be drawn from the impressive results that models have obtained in recent years as deep learning methodologies continue to develop and show improvements.